%% file: NLEguide.tex
\documentclass{nle}

\usepackage{dirtytalk}
\usepackage[most]{tcolorbox}
\usepackage{amsmath}
\usepackage{subcaption}
\usepackage{setspace} 
\usepackage{blindtext}
\usepackage{hyperref}
\usepackage{csquotes}
\usepackage{booktabs}
\usepackage{caption}
\usepackage{longtable}
\usepackage{spverbatim}
\usepackage{amsfonts}
\usepackage{graphicx}
\usepackage{natbib}
\ifpdf%
\usepackage{epstopdf}%
\else%
\fi
\usepackage{multirow}
\usepackage[switch, modulo]{lineno}

\begin{document}
\label{firstpage}

\righttitle{J Pereira et al.}

\papertitle{Article}

\jnlPage{1}{00}
\jnlDoiYr{2023}

\title{Predictive Authoring for Brazilian Portuguese Augmentative and Alternative Communication}

\begin{authgrp}
\author{Jayr Pereira$^{1,2,*}$, Rodrigo Nogueira$^{2}$, Cleber Zanchettin$^1$ and Robson Fidalgo$^1$}
\affiliation{$^1$Centro de Informática, Universidade Federal de Pernambuco, Av. Jornalista Aníbal Fernandes, s/n, CEP 50.670-901, Recife-PE,  Brazil}
\affiliation{$^2$NeuralMind, Brazil}
\affiliation{$^*$Corresponding author. E-mail: jap2@cin.ufpe.br}
\end{authgrp}

\history{Competing interests: The author(s) declare none.}

\begin{abstract}

Individuals with complex communication needs (CCN) often rely on augmentative and alternative communication (AAC) systems to have conversations and communique their wants. Such systems allow message authoring by arranging pictograms in sequence. However, the difficulty of finding the desired item to complete a sentence can increase as the user's vocabulary increases. This paper proposes using BERTimbau, a Brazilian Portuguese version of BERT, for pictogram prediction in AAC systems. To finetune BERTimbau, we constructed an AAC corpus for Brazilian Portuguese to use as a training corpus. We tested different approaches to representing a pictogram for prediction: as a word (using pictogram captions), as a concept (using a dictionary definition), and as a set of synonyms (using related terms). We also evaluated the usage of images for pictogram prediction. The results demonstrate that using embeddings computed from the pictograms' caption, synonyms, or definitions have a similar performance. Using synonyms leads to lower perplexity, but using captions leads to the highest accuracies. This paper provides insight into how to represent a pictogram for prediction using a BERT-like model and the potential of using images for pictogram prediction.

\end{abstract}

\maketitle

\input{sections/introduction.tex}
\input{sections/background.tex}

\input{sections/method.tex}
\input{sections/results.tex}
\input{sections/conclusions.tex}

\section*{Acknowledgement}

This research was supported by the Coordenação de Aperfeiçoamento de Pessoal de Nível Superior - Brazil (CAPES). Grant code: [88887.481522/2020-00]. The pictographic symbols used are the property of the Government of Aragón and have been created by Sergio Palao for ARASAAC (http://www.arasaac.org), that distributes them under Creative Commons License BY-NC-SA.

\label{lastpage}
\bibliographystyle{nlelike}
\bibliography{refs}

\end{document}

%% file: sections/introduction.tex
\section{Introduction} \label{sec:introduction}

Augmentative and Alternative Communication (AAC) systems are tools used by people with Complex Communication Needs (CCN) (e.g., people with Down's syndrome, autism spectrum disorders, intellectual disability, cerebral palsy, developmental apraxia of speech, or aphasia) to compensate for the difficulties faced in their daily communication \citep{beukelman1998augmentative,asha2019defintion}.
According to \cite{beukelman1998augmentative}, approximately 97 million people worldwide may benefit from AAC. 
These people constitute a heterogeneous population regarding diagnosis, age, location, communication modality, and extent of AAC use \citep{asha2019defintion}. 
They generally have limitations on gestures, and oral and written communication, causing functional communication and socialization problems. 
AAC users include more than just people with CCN. It also includes children at risk for speech development, individuals who require AAC to supplement and clarify their speech or support comprehension (e.g., those with degenerative cognitive and linguistic disorders such as Alzheimer's disease), and those with temporary conditions \citep{beukelman1998augmentative}.

AAC tools are often categorized into \textit{low-tech} (e.g., papercraft cards) and \textit{high-tech} (e.g., speech-generating devices).
Low-tech AAC systems like papercraft cards or picture exchange communication systems (PECS) offer people with CCN a simple and tangible way to express themselves. These systems involve selecting various images or objects representing words or concepts, allowing users to construct sentences and visually express their thoughts. They are instrumental when power sources or sophisticated digital technology are not readily available or manageable. While these systems might not be as sophisticated as their high-tech counterparts, they can provide a foundation for language development and are often highly portable and easy to use. On the other hand, high-tech AAC systems rely on more complex devices such as speech-generating devices, tablets with dedicated apps, or computer software that can facilitate communication. Such devices typically combine text, symbols, and/or voice output. 

\begin{figure}[t]
\centering
\tcbox{\includegraphics[width=13cm]{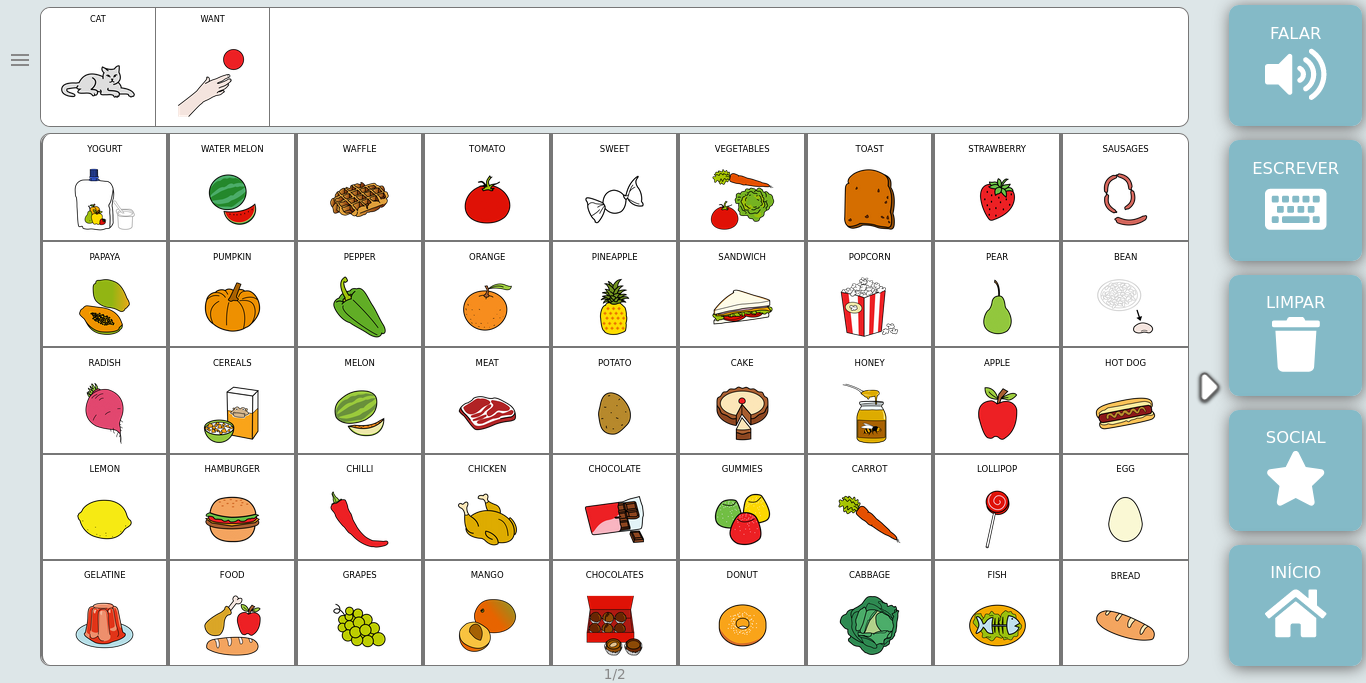}}
\caption{Example of a high-tech AAC system using communication cards with ARASAAC pictograms. The screenshot depicts the interface of the reaact.com.br tool, where the user can easily select communication cards from the content grid (large bottom rectangle) and arrange them sequentially to construct sentences (e.g., \textit{cat wants}). Additional functionalities are accessible through the buttons located in the right sidebar, enabling utilities such as text-to-speech functionality provided by the voice synthesizer.}
\label{fig:pca_example}
\end{figure}

High-tech AAC systems help users to express feelings and opinions, develop understanding, reduce frustration in trying to communicate, and help to communicate preferences and choices \citep{beukelman1998augmentative}. 
Such systems have been gaining ground in recent years. The advent of mobile devices such as iPad, iPhone, and Android smartphones and tablets facilitated the release of low-cost systems \citep{Lorah2018,Lorah2022}. By searching in the Apple App Store and Google Play Store for \say{alternative communication}, one can find a variety of applications for AAC. Most apps promote communication using pictograms, similar to the one shown in Figure \ref{fig:pca_example}. Studies have demonstrated the positive effect of these devices' usage by people with CCN \citep{Holyfield2022,hughes2022direct}. \cite{Holyfield2022} showed that using high-tech AAC is more pleasant for children with multiple disabilities compared to \textit{low-tech} devices. Besides, they suggest that using high-tech systems may be more efficient.
These systems allow users to construct sentences by selecting communication cards (a.k.a. pictograms) from a grid and arranging them sequentially. Figure \ref{fig:pca_example} presents an example of a high-tech AAC system with a content grid (large bottom rectangle) and a sentence area (tiny top rectangle), where cards are arranged in sequence. 

Recent advancements have significantly enhanced the integration of AI into AAC systems. As \cite{elsahar2019augmentative} points out, incorporating AI into AAC systems can lead to increased accessibility to high-tech devices, faster output generation, and improved customization and adaptability of AAC interfaces. The potential benefits of AI in AAC systems are also highlighted by \cite{Sennott2019}, who explicitly mentions the application of Natural Language Processing (NLP) techniques for tasks such as word and message prediction, automated storytelling, voice recognition, and text expansion. The use of AI in AAC systems opens up possibilities for assisting in creating grammatically correct, semantically meaningful, and comprehensive messages. For instance, predictive models can be used to aid in message authoring \citep{pereira2020semantic,PEREIRA2022pictobert,Hervas2020,Dudy2018,garcia2016evaluating,garcia2015context}. These studies are driven by the need for AAC systems to facilitate the construction of meaningful and grammatically correct sentences \citep{franco2018towards}. Moreover, predictive models in AAC can offer numerous benefits to users \citep{beukelman1998augmentative}, such as:
1) reducing the number of selections needed to construct a sentence, thereby decreasing the communication effort; 
2) providing spelling support for users who struggle with accurate spelling; 
3) offering grammatical support; and 
4) increasing the communication rate (words per minute).

In a recent survey, \cite{pereira2022mapping} listed eight studies proposing pictogram prediction methods in AAC. The survey's results indicate that the methods used for prediction have changed over time, ranging from knowledge databases to statistical language models. \cite{PEREIRA2022pictobert} demonstrated that fine-tuning BERT for pictogram prediction leads to better performance and generalization than n-gram language models and knowledge databases. However, the proposed model's ability to adapt to different users or user groups' needs, using it for languages other than English, is still problematic. The main difficulty is the lack of corpora to be used for training. Previous works used conversational natural language corpora adapted for AAC \citep{Dudy2018,PEREIRA2022pictobert}.

This paper proposes using BERT for pictogram prediction in Brazilian Portuguese. It involves constructing and using an AAC corpus to finetune BERTimbau \citep{souza2020bertimbau}, a Brazilian Portuguese version of BERT. For corpus construction, we first collect AAC-like sentences constructed by AAC practitioners; then, we use GPT-3 \cite{brown2020gpt3} to generate similar synthetic sentences; finally, we convert the natural language sentences into pictogram-based sentences. For BERTimbau finetuning, we adapted the model by changing its vocabulary and embedding layer to handle the vocabulary present in the generated synthetic corpus. We tested the different approaches found in the literature on how to represent a pictogram in pictogram prediction: as a word (using pictogram captions), as a concept (using a dictionary definition), and as a set of synonyms (using related words). With these tests, we aim to answer the following question \textit{What is the best way to represent a pictogram for prediction using a BERT-like model?}. Besides, considering that a pictogram is a visual support for communication in AAC systems, we assessed the usage of images for pictogram prediction. The goal is to answer the question \textit{Can image representations increase the quality of pictogram prediction using a BERT-like model?}

We evaluated the performance of model variations in terms of perplexity and top-$n$ accuracy. We use $n \in \{1,9,18,25,36\}$ to simulate the different grid sizes an AAC system can have. The results demonstrate that using embeddings computed from the pictograms' caption, synonyms, or definitions have a similar performance. Using synonyms leads to lower perplexity, but using captions leads to the highest accuracies. This way, choosing a method to implement in an AAC system is a design decision. A lower perplexity indicates that the model can generalize unseen data well. However, using synonyms requires the preexistence of a database of synonyms. Using only captions can cause problems when the vocabulary has many pictograms for the same word. An alternative to solving this is using the pictogram definition, as in a dictionary. Previous studies demonstrated that a pictogram is better represented by a dictionary concept \citep{PEREIRA2022pictobert,schwab2020providing}. However, the definition-based finetuning did not outperform the caption- and synonyms-based in our experiments. Using images for compute embeddings requires more training data and time, for the vectorial space differs from the BERTimbau input embeddings. The code for these experiments is available online\footnote{\url{https://github.com/jayralencar/pictogram_prediction_pt}}.

The findings of this paper hold valuable implications for researchers, practitioners, and developers engaged in AAC systems, particularly those aiming to incorporate communication card prediction into their systems. The target audience for such systems typically comprises children with complex communication needs who face challenges in conventional writing or utilizing a traditional keyboard, such as QWERTY, for communication purposes. It is important to note that the intended users of these systems may or may not be literate. In the case of literate children, cognitive deficits may hinder their ability to effectively use written language, making AAC systems a supportive tool for communication. Alternatively, AAC is an alternative resource for non-literate children, as it relies on a graphical system rather than conventional writing. By leveraging the insights and methodologies presented in this paper, researchers, practitioners, and developers can enhance the design and functionality of AAC systems, ultimately enabling effective and efficient communication for this target audience.

This paper is organized as follows: in Section \ref{sec:background}, we present the theoretical information that is this work's basis; in Section \ref{sec:method}, we present the proposed method for finetuning BERTimbau and experimental details; in Section \ref{sec:results}, we present our results; and, finally, in Section \ref{sec:conclusions}, we present the conclusions.


%% file: sections/background.tex
\section{Background} \label{sec:background}

\subsection{Language Modeling} \label{sec:background_language model}

A language model assigns probabilities to sequences of words \citep{jurafsky2019speech}. Consider the sentence \say{Brazil is a beautiful \rule{1cm}{0.15mm}} and ask what is the best word to complete it. Most people will choose words such as \say{country}, \say{place}, or \say{nation}, for they are the most probable among those that occur in natural language texts. This human decision is so natural that we do not think about how it happens. However, for language models, deciding which word to use to complete a sentence depends on the probabilities learned from a training corpus. For example, for an n-gram language model, the most probable word is the one that occurs most frequently after the word \say{beautiful} in the training corpus. The same model can also assign a probability to an entire sentence and predict that the sentence \say{Brazil is a beautiful country} has a higher probability of appearing in a text corpus than the same words in a different order (e.g., \say{is country beautiful Brazil a}).

An n-gram language model is the simplest model that assigns probabilities to sequences of words \citep{jurafsky2019speech}. The aim is to predict the next word based on the $n-1$ preceding words. The model uses relative frequency counts to estimate the probability of each word in a vocabulary $V$ to be the next in the sequence $h$.
Given a large text corpus, one counts the number of times the sequence $h$ is followed by the word $w \in V$. This way, in a bigram model ($n=2$), the probability of the word \say{country} completing the sequence \say{Brazil is a beautiful \rule{1cm}{0.15mm}} can be simplified to: 

\begin{equation}
    \label{eq:bigram}
    P(country|Brazil\ is\ a\ beautiful) = \frac{C(beautiful\ country)}{C(beautiful)}, 
\end{equation}

\noindent where $C$ is the function that counts the occurrence of words or sequences in the corpus. Since this is a bigram model, only the last preceding word is considered in the equation, which can be simplified to $P(country|beautiful)$ or $P(w_n|w_{n-1})$.

The probability of an entire sequence can be estimated using the chain rule:

\begin{equation}
    \label{eq:chain_rule}
      \begin{aligned}
        P(w_{1:n})  & = P(w_1)P(w_2|w_1)P(w_3|w_{1:2})...P(w_n|w_{1:n-1}) \\
                    & = \prod_{k=1}^{n} P(w_k|w_{1:k-1})
    \end{aligned}
\end{equation}

The assumption that the probability of the next word depends only on the previous word is called the Markov assumption \citep{jurafsky2019speech}. Markov models assume that it is possible to predict the probability of a future unit (e.g., next word) by looking only at the current state (e.g., last preceding word). However, language is a continuous input stream highly affected by the writer/speaker's creativity, vocabulary, language development level, etc. Suppose one asks two people to describe the same scene from a picture in a single sentence. In that case, there is a probability of both constructing sentences with similar semantics but using different words or ordering them differently. Besides, in a written text, the occurrence of a specific word may depend not only on the $n-1$ preceding words but on the entire context, which can be the sentence, the paragraph, or all of the text. Still, n-gram models produce strong results for relatively small corpora and have been the dominant language model approach for decades \citep{yoav2017neural}. 

Among the language models that do not make the Markov assumption, we can highlight those based on recurrent neural networks (RNN) \citep{elman1990rnn} and the Transformers architecture \citep{vaswani2017attention}. Both may rely on word embeddings for feature extraction.

\subsubsection{Word Embeddings} \label{sec:word_embeddings}

Word embeddings is a method to represent words using real-valued vectors to encode their meaning, assuming that words with similar meanings may be closer to each other in the vector space \citep{jurafsky2019speech}. 
\cite{mikolov2013word2vec} proposed the skip-gram model (a.k.a. \textit{word2vec}), which learns high-quality vector representations of words from large amounts of text. The quality of the learned vectors allows similarity calculations between words and even operations such as $King - Man + Woman = Queen$, or $Madrid - Spain + France = Paris$. This means that by subtracting the vector of the word \textit{Man} from the vector of the word \textit{King} and summing it with the vector of the word \textit{Woman}, the resulting vector is closer to the vector of the word \textit{Queen} than any other vector \citep{mikolov2013linguistic,mikolov2013word2vec}.
These vectors can also capture synonymy with quality, for words with similar meanings might have similar vector representations.

The Skip-gram model's training objective is to find word vectors useful for predicting the surrounding words in a sequence or a document \citep{mikolotov2013distributed}. This way, the model is trained using a self-supervised approach, which avoids the need for any hand-labeled supervision signal. 
Given a sequence of words $w_1,w_2,...,w_n$, the model attempts to maximize the average log probability calculated according to Equation \ref{eq:skipgram_log_probability}), where $c$ is the training context size of words that are surrounding the center word $w_t$. A large $c$ results in more training examples and can lead to a high accuracy but may require more training time \cite{mikolotov2013distributed}. The basic Skip-gram formulation defines $P(w_{t+j} |w_t)$ using the softmax function, as in Equation \ref{eq:skipgram_softmax}, where $v_w$ and $v^{'}_{w}$ are the input and output vectors of $w$, and $W$ is the vocabulary size. This formulation is impractical for the cost of computing the gradient of $log P(w_O|w_I)$ is proportional to the vocabulary size, which can be large. \cite{mikolotov2013distributed} suggests using the hierarchical softmax \cite{morin2005hierarchical} as an efficient approximation of the full softmax. This way, the neural network behind skip-gram learns the best vector representation for each word in a vocabulary. The final model output is a dictionary with $\{word: vector\}$ pairs.

\begin{equation}
    \label{eq:skipgram_log_probability}
    \frac{1}{n}\sum_{t=1}^{n} \sum_{-c \leq j \leq  c, j \neq 0} log P(w_{t+j}|w_t)
\end{equation}

\begin{equation}
    \label{eq:skipgram_softmax}
    P(w_O|w_I) = \frac{
        \exp(
            {v^{'}_{w_{O}}}^{\top} v_{w_{I}}
        )
    }
    {
        \sum_{w=1}^{W} \exp({v^{'}_{w}}^{\top} v_{w_{I}})
    }
\end{equation}

There is a set of other word embedding approaches with the same aim: to provide vector representation to words. 
We can classify skip-gram as a model that provides static embeddings, for the representation of a word will be the same indifferently of the context it occurs. For example, the word \textit{bat} has a different meaning in the sentences \textit{He can't bat the ball} and \textit{Batman dress like a bat}. However, in a static word embedding model, it has the same vector. 
The transformers architecture \citep{vaswani2017attention} overcomes this problem by adding context to the embeddings.

\subsubsection{Transformers} \label{sec:background_transformers}

The Transformers architecture, introduced by \cite{vaswani2017attention}, is a neural network model that operates solely on self-attention mechanisms to compute input and output representations. This innovative approach allows for efficient and effective sequential data processing in various natural language tasks. 
Self-attention allows a Transformer to extract and use information from arbitrarily large contexts without passing it through intermediate recurrent connections as in RNNs \citep{jurafsky2019speech}. A self-attention layer maps the input sequences to output sequences of the same length. While processing the input, the model can access all the inputs, including the one in consideration. However, it has no access to information concerning inputs beyond the current one. The self-attention allows the model to relate different positions of a single sequence to compute the representation sequences' items. By doing so, an attention-based approach compares an item of interest to a collection of other items to reveal their relevance in the context (or sequence) \citep{jurafsky2019speech}. Given a sequence input, a transformer produces an output distribution over the entire vocabulary for language modeling. The most famous language models based on transformers architecture are GPT series \citep{radford2018improving,radford2019language,brown2020gpt3} and BERT \citep{devlin-etal-2019-bert}.

GPT \citep{radford2018improving,radford2019language,brown2020gpt3} is an auto-regressive generative language model standing for Generative Pre-trained Transformers. This model uses the Transformers architecture to learn word representation that transfers with little adaptation to a wide range of tasks \cite{radford2018improving}. The main task is to predict the next word in a given sequence and then learn the best vectorial word representations. These representations perform downstream tasks like sentiment analysis, machine translation, etc. 
The most recent version of the series is the GPT-3 \citep{brown2020gpt3}, demonstrating that Llanguage models are few-shot learners. This model and its rivals (e.g., Google PaLM model \citep{palm2022}, and DeepMind GOPHER \citep{gopher}) promoted a revolution in most of the NLP-related tasks for not huge amounts of annotated data are necessary to a downstream task. GPT-3 was trained with 100 times more data than its predecessor GPT-2. A large amount of training data and the high number of used parameters make GPT-3 powerful in performing on-the-fly tasks that were never explicitly trained. Among these tasks, we can cite machine translation, math operations, writing code, etc. 

BERT is a language representation model that stands for Bidirectional Encoder Representations from Transformers \citep{devlin-etal-2019-bert}. This model uses the attention mechanism \citep{vaswani2017attention} to learn contextual relations between tokens (words or sub-words) in unlabeled texts by jointly conditioning on both left and right contexts in all model layers. Unlike directional models, which process the input in sequence (left-to-right or right-to-left), BERT processes the entire sequence simultaneously. Thus, it allows the model to learn the word's context based on all neighborhoods, left and right. To do this, the model performs masked language modeling (MLM). During training, the data generator randomly chooses 15\% of the token positions for prediction. For example, if the $i$-th token is chosen, it is replaced with (1) the $[MASK]$ token 80\% of the time, (2) a random token 10\% of the time, or (3) the unchanged $i$-th token 10\% of the time.
The model attempts to predict the $i$-th token based on the contextual information provided by the non-masked, generating a contextualized representation for each.

\subsubsection{Evaluating Language Models} \label{sec:perplexity}

One approach to assess the quality of a language model is to implement it in an application and evaluate its performance improvement, known as an extrinsic evaluation \citep{jurafsky2019speech}. However, this requires creating a complete system that uses the $n$ models being evaluated, which can be both time-consuming and computationally expensive. For example, if two models for pictogram prediction were being compared, the models would need to be trained, two AAC boards using each model would need to be created, and a metric related to communication would need to be measured. This process can require a lot of human and computational resources, making it difficult or even impossible to complete. On the other hand, an intrinsic evaluation metric assesses the quality of a model without taking any application into account \citep{jurafsky2019speech}.

Perplexity ($PP$ or $ppl$), an intrinsic evaluation metric, offers a quick and easy way to compare language models. It only requires a training and test dataset, making it a fast and low-resource experiment. Moreover, recent studies suggest that perplexity is correlated with the human judgment of sentences generated by language models \citep{shen2017estimation,crook2017sequence,adiwardana2020towards}.
The perplexity of a language model is a measure of how well it comprehends language. It is calculated by taking the inverse probability of the test set, divided by the number of unique words in the vocabulary \citep{jurafsky2019speech}. A low perplexity indicates that the test set is not too surprising for the model, meaning it understands the language well. As an example, let's say the test set is $W = w_1, w_2, ..., w_N$:

\begin{equation}
    PP(W) = P(w_1, w_2, ..., w_N)^{-\frac{1}{N}} = \sqrt[N]{\frac{1}{P(w_1, w_2, ..., w_N)}}
\end{equation}

The probability of $W$ can be expanded with the chain rule:

\begin{equation} \label{eq:inverse}
    PP(W) = \sqrt[N]{\prod_{i=1}^{N}\frac{1}{P(w_i|w_1,...,w_i-1)}}
\end{equation}

Where $P(w_i|w_1,...,w_i-1)$ is the probability of the $i$-th token given the previous $i-1$ (i.e., the context). Thus, for a bigram model:

\begin{equation}
    PP(W) = \sqrt[N]{\prod_{i=1}^{N}\frac{1}{P(w_i|w_i-1)}}
\end{equation}

Notice that because of the inverse in Equation \ref{eq:inverse}, the higher the conditional probability of the word sequence, the lower the perplexity.

We can calculate perplexity by exponentiating the cross-entropy. This gives us an estimate of the average number of words required to encode a given sequence of words using $H(W)$.

\begin{equation} \label{eq:perplexity}
    PP(W) = 2^{H(W)} = 2 ^{-\frac{1}{N} log_2 P(w_1, w_2,...,w_N)}
\end{equation}

BERT MLM does not directly compute perplexity since the cross-entropy is only calculated for masked tokens. But BERT does give the probability of a sentence from test sets by assigning the probability of each word when no masked token is inputted into the model. We can then use this sentence probability to calculate the cross-entropy and the perplexity.

\subsection{Pictogram Prediction in AAC} \label{sec:background_pictogram_prediction}

AAC employs a variety of tools and techniques to aid the communication of individuals with CCN. In the context of \textit{high-tech} AAC, pictographic images on communication cards serve as visual aids for the user, providing meaning to the words in their vocabulary. These pictograph systems benefit illiterate individuals due to age or disability, enabling communication for those with low cognitive levels or at very early stages \citep{arassac_2019}. Numerous online databases offer a wealth of pictograms. However, there is no dataset as extensive as ARASAAC \cite{arassac_2019} for Brazilian Portuguese, making it the best alternative available. It provides access to over 30 thousand pictograms. Many of the available \textit{high-tech} AAC systems arrange the pictograms in grids, as depicted in Figure \ref{fig:pca_example}. The organization of the vocabulary is tailored to the user's needs and preferences. Some may opt to categorize the cards, while others may prefer multiple pages. Nevertheless, these systems must facilitate card selection for sentence construction \citep{franco2018towards}.

Among the strategies to facilitate card selection in \textit{high-tech} AAC systems, we can mention four as the main ones: 
1) vocabulary organization -- organize the cards meaningfully to facilitate searching (e.g., taxonomic organization);
2) color coding systems usage -- use some color coding system to label cards, such as the Fitzgerald Keys \citep{fitzgerald1949straight,mcdonald1973communication} or Colourful Semantics \cite{bryan1997colourful};
3) motor planning strategies -- use consistent motor patterns to facilitate card findings throughout motor memory (e.g., using the LAMP protocol \citep{halloran2006lamp}); and
4) the usage of predictive models -- predict the next cards suitable to complete sentences in construction.
Predictive models can be used in addition to the other strategies to further refine the search for communication cards. Besides, the benefits of using prediction techniques in AAC include \citep{beukelman1998augmentative}:
1) reducing the number of selections required to construct a sentence, thereby decreasing the effort for individuals; 
2) providing spelling support for users who cannot accurately spell words; 
3) providing grammatical support; and 
4) increasing communication rate (i.e., words per minute). 

Communication card prediction in AAC assumes a controlled vocabulary containing the cards used in the user's daily communication. The language model assigns the probability of each vocabulary item being the next in an in-construction sentence. Recent studies used different models to perform this role. The most common are based on knowledge bases \citep{pereira2022mapping}. Such models may allow using semantic scripts like the Colourful Semantics \citep{bryan1997colourful,pereira2021caregivers} as support for sentence construction. However, they generally rely on complex construction pipelines, which require reprocessing for vocabulary or knowledge updates. Training a statistical language model might be an alternative. Some other proposals use n-gram \citep{garcia2016evaluating,Hervas2020} or neural network \citep{Dudy2018,PEREIRA2022pictobert} models. The literature suggests that neural networks based language models may perform better than n-gram models \citep{yoav2017neural}. However, they may require more computational resources for training and serving, making their deployment difficult in production.

Choosing a pictogram prediction model may involve practical questions like computational resources, deployment, etc. However, it also involves conceptual decisions. An example is the decision of what a pictogram is. Simply, a pictogram is a graphic symbol representing an object or concept. It is usual to see pictograms on traffic signs, for example. In AAC, a pictogram is generally associated with a caption with the word or expression it represents. The pair pictogram-caption forms the communication card, which the user selects and organizes to constitute a sentence. Some pictogram prediction approaches feed their models only with the captions \citep{garcia2016evaluating,Hervas2020,saturno2015an}, considering the task as a word prediction task. Other studies consider an AAC pictogram as a concept that links the graphical representation and the caption \citep{pereira2020semantic,PEREIRA2022pictobert,Dudy2018,martinez2015semantic}. \cite{schwab2020providing} consider that a concept from a dictionary better represents a pictogram (e.g., person: a human being). They associated the ARASAAC pictograms with synsets (i.e., concepts) from the Princenton WordNET\footnote{\url{https://wordnet.princeton.edu/}}, a lexical database for English.

For prediction, using concepts may be more meaningful because of polysemic words. For example, the English word \say{bat} can have many meanings (e.g., \say{nocturnal mouselike mammal} or \say{a club used for hitting a ball}) and, similarly, many related pictograms in a given vocabulary. The way how to do this varies among approaches. \cite{Dudy2018} grouped the words related to each pictogram and calculated embeddings to feed their LSTM model. While \cite{PEREIRA2022pictobert} associated each pictogram to a WordNET synset and used the vectors calculated by \cite{scarlini2020ares} for each synset as inputs of their BERT-based model.

Although there are proposals for predicting pictograms in AAC, a pictogram-based corpus is not available. \cite{PEREIRA2022pictobert} and \cite{Dudy2018} used corpora in natural language adapted to the task. \cite{PEREIRA2022pictobert} proposed SemCHILDES, which consists of part of the Child Language Data Exchange System (CHILDES) \citep{macwhinney2014childes} dataset annotated with word senses. The corpora in CHILDES have transcribed conversations between children and parents, therapists, or teachers. The conversational nature and the public audience may make it suitable for pictogram prediction in AAC. However, following \cite{PEREIRA2022pictobert}'s pipeline, the corpus requires some pre-processing steps.
In comparison, \cite{Dudy2018} used an adapted version of SubtlexUS \citep{brysbaert2009moving}, a corpus of subtitles from movies and television. The authors used the corpus as a proxy for AAC due to its spontaneous speech. \cite{Vertanen2013} proposed a corpus with everyday conversation communications. The corpus has natural language sentences produced by workers from a crowdsourcing site. Thus, it is not properly an AAC corpus.

%% file: sections/method.tex
\section{Method} \label{sec:method}

\begin{figure}[t]
\centering
\includegraphics[width=\textwidth]{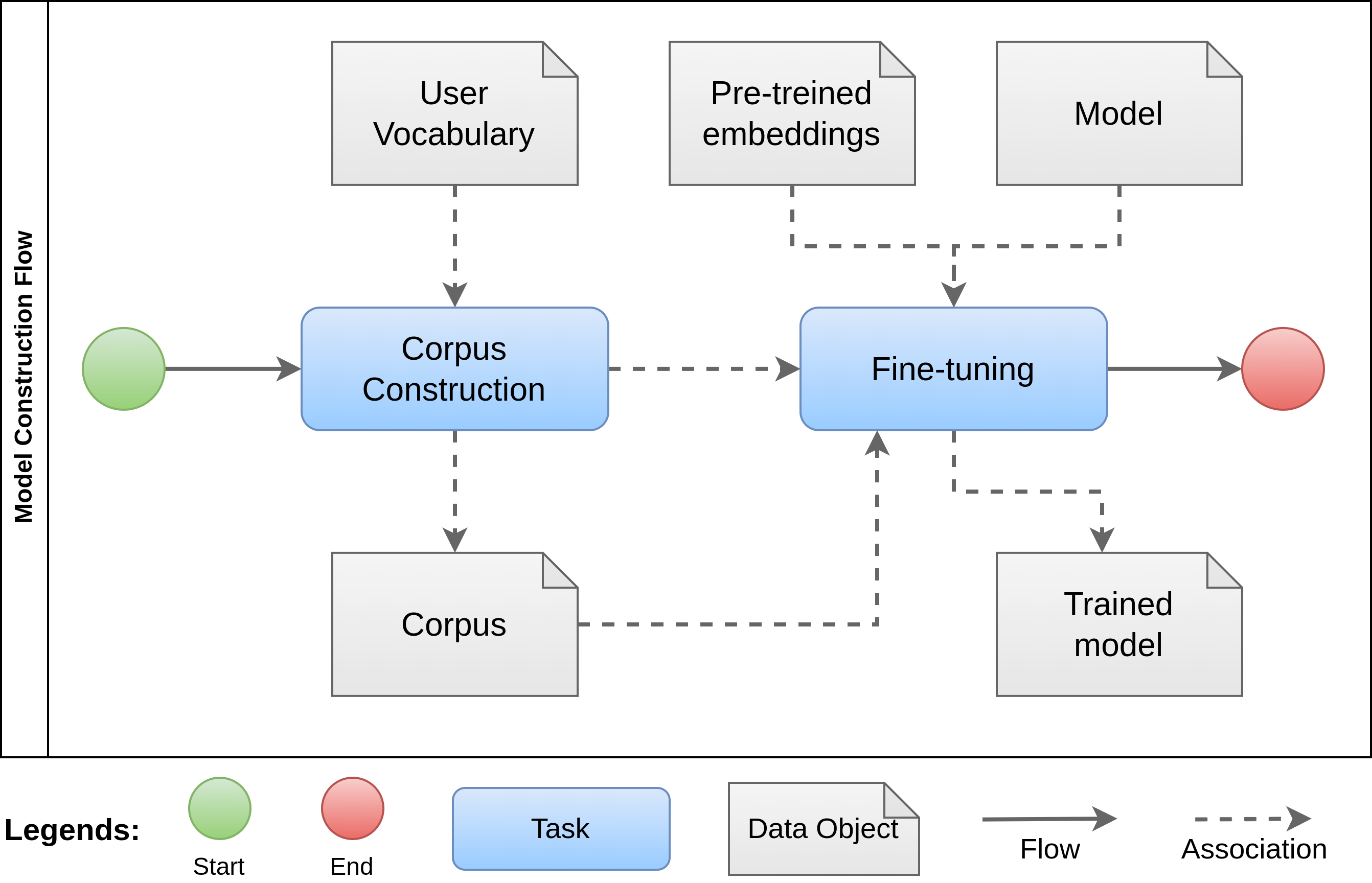}
\caption{Flow chart for model construction.}
\label{fig:constructing_flow}
\end{figure}

This section details our proposed method for fine-tuning BERTimbau for pictogram prediction in high-tech AAC systems. 
Figure \ref{fig:constructing_flow} illustrates the method flow. The three main inputs are a controlled vocabulary, a pre-trained embedding matrix, and a pre-trained transformer. 
We detail inputs in Section \ref{sec:method_inputs}. 
The two main steps are 1) corpus construction (cf. Section \ref{sec:corpus_construction}), and 2) Model fine-tuning (cf. Section \ref{sec:finetuning}). The main output of this method is the fine-tuned model, but we also consider the constructed corpus as a relevant output.


\subsection{Inputs} \label{sec:method_inputs}

The three input resources for our method are 1) a pre-trained BERT, 2) a controlled vocabulary, and 3) a pre-trained embedding matrix. As an input model, we used BERTimbau \citep{souza2020bertimbau}, a Brazilian Portuguese version of BERT. 
As a controlled vocabulary, we consider a list of communication cards, each consisting of 1) a pictogram (or picture) and 2) a caption with a word or a multi-word expression. It is common in the AAC field to have pre-defined vocabularies aimed at different contexts, activities, etc. An example is Project-Core\footnote{\url{http://www.project-core.com/communication-systems/}}, which defines a list of 36 symbols as sufficient for a universal core communication. 
Our experiments use the list of pictograms for Brazilian Portuguese from the ARASAAC dataset. There are 12785 pictograms related to words and multi-word expressions (e.g., \say{café da manhã}, i.e., breakfast) \footnote{Available at \url{https://api.arasaac.org/api/pictograms/all/br}. Accessed on December 21, 2022.}.

As mentioned in Section \ref{sec:word_embeddings}, word embeddings are real-valued vectors used to represent words. In our experiments, we extract embeddings from four sources:
1) the pictogram caption (i.e., word or expression);
2) the pictogram caption synonyms;
3) the pictogram glossary definition from ARASAAC; and
4) the pictogram image.
For the caption embeddings, we use the input vectors from BERTimbau as a basis. Formally, given a vocabulary $V$ composed of words and MWEs $(w_1, ..., w_n)$, the BERTimbau original embedding $B \in \mathbb{R}^{h \times D_b}$, where $h$ is the size of the hidden state and $D_b$ is the BERTimbau vocabulary size, and given a new embeddings matrix $P \in  \mathbb{R}^{h \times D_v}$, where $D_v = |V|$, for each token $t_i$ in $V$, populate $P$ with the $t_i$ embeddings from $B$. 
For MWEs, the embeddings of each token are extracted from BERTimbau's embeddings layer to a matrix $E \in \mathbb{R}^{h \times n}$, where $h$ is the dimensionality of the embedding (the same of hidden states size), and $n$ is the number of tokens in the expression. We use the mean vector $\overline{E}$ as the expression's embedding representation. We use an approach similar to \cite{Dudy2018} for caption synonyms. First, we search in ARASAAC for the list of keywords for each pictogram. The pictogram representation is the average of the embeddings of its keywords.

For generating embeddings from pictogram definition, we applied two methods. Both methods use the definitions from ARASAAC concatenated with keywords. A pictogram in ARASAAC has a list of keywords, which have a definition each. We concatenate this list as $keyword_i || definition_i || ... || keyword_n || definition_n$. The first extraction method considers the mean vector of the definition extracted from $B$ (i.e., BERTimbau input embeddings). 
The second method uses the BERTimbau last encoders layer outputs for the $[CLS]$ token. \footnote{BERT tokenizer adds the $[CLS]$ token at the beginning of the processed sentences. This token output representation is generally used as input for classification models.} 
We also computed representations from pictogram images using a Vision Transformer (ViT). We used a ViT model pre-trained on ImageNet-21k (14 million images, 21,843 classes) and fine-tuned on ImageNet 2012 (1 million images, 1,000 classes) \citep{Dosovitskiy2020animage}\footnote{Available at \url{https://huggingface.co/google/vit-base-patch16-224}}.  



\subsection{Corpus Construction} \label{sec:corpus_construction}

This section presents the method for constructing our corpus. Our method consists of augmenting a set of sentences constructed by AAC practitioners. For this, we applied a four-step pipeline: 
1) collection of sentences (cf. Section \ref{sec:sentence_collection});
2) data augmentation (cf. Section \ref{sec:corpus_augmentation});
3) data cleaning (cf. Section \ref{sec:data_cleaning}); and
4) text-to-pictogram transformation (cf. Section \ref{sec:text_to_pictogram}).
Section \ref{sec:corpus_pt_analysis} presents an analysis of the corpus's main features.

\subsubsection{Collection of Sentences} \label{sec:sentence_collection}

\begin{figure*}[t]
    \centering
    \begin{subfigure}[b]{0.5\textwidth}
        \centering
        \includegraphics[width=\textwidth]{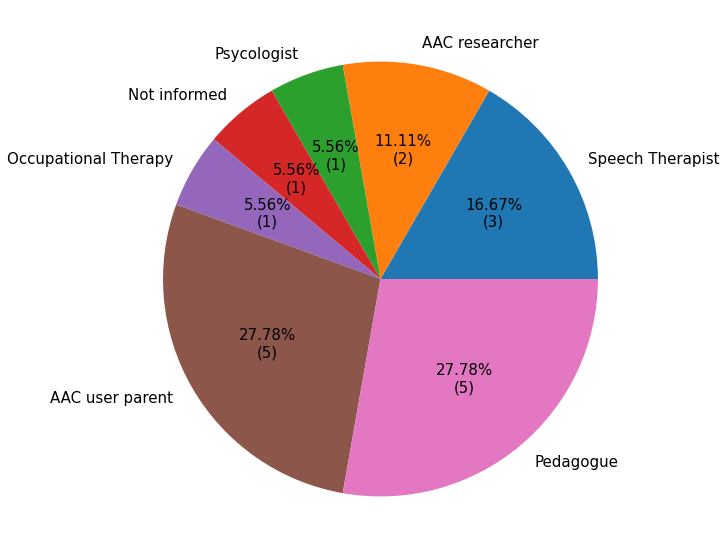}
        \caption{Participant profile.}
        \label{fig:participant_profile}
    \end{subfigure}
    \hfill
    \begin{subfigure}[b]{0.45\textwidth}
        \centering
        \includegraphics[width=\textwidth]{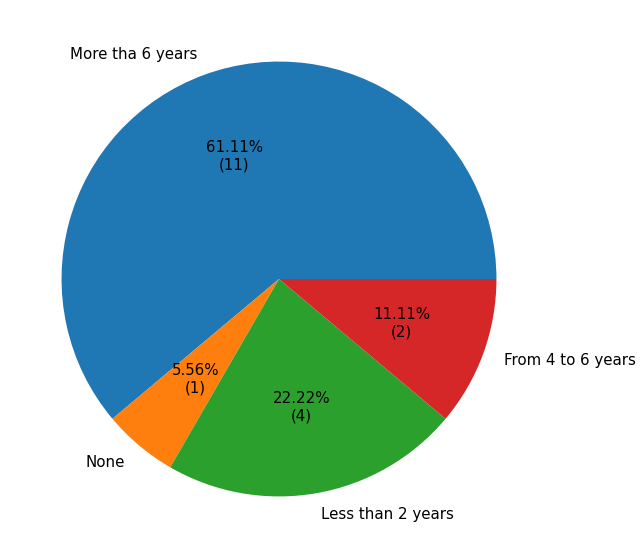}
        \caption{Participant experience with AAC.}
        \label{fig:participant_experience}
    \end{subfigure}
    \caption{Sentence collection participants summary.}
    \label{fig:participant_summary}
\end{figure*}

For collection of sentences, we invited speech therapists, psychologists, and parents of children with CCN to inform the sentences they consider the most commonly constructed in different contexts using high-tech AAC. 
We make an online questionnaire available and send it to groups of people interested in AAC. In addition, we invited experts who had participated in other studies that we had previously conducted.
Seventeen individuals participated in this study. Figure \ref{fig:participant_summary} presents a summary of the participants. Most have used AAC with their children or patients for more than six years. That is, they have vast experience in using such tools. Besides, we had participants from at least six different fields, who may observe the AAC usage from various points of view. 

Each participant answered a questionnaire with six questions asking them to construct sentences. The first four questions required sentences about home, school, kitchen, and leisure contexts. The fifth question asked the participant to construct sentences that describe events free of context (e.g., I ate eggs at breakfast today). The last question asked them to construct sentences free of context that they consider essential for AAC. With this procedure, we collected a total of 667 unique sentences.

\subsubsection{Data Augmentation} \label{sec:corpus_augmentation}

The data augmentation step aims to generate sentences similar to those constructed by AAC practitioners, which we now refer to as human-composed. The generated sentences must be similar regarding used words (vocabulary) and sentence structure (semantics and syntax).
We used GPT-3 \cite{brown2020language}\footnote{We used \texttt{text-davinci-002} available via the OpenAI API.} with a few-shot learning approach. 
We provide some examples to GPT-3 in the form of text prompts and ask it to produce new similar examples by completing our prompts.
We used two approaches: 1) using the human-composed sentences as examples; and 2) using a controlled vocabulary as a basis. We explain each approach in more detail below.


We used the human-composed sentences as few-shot examples in the GPT-3 prompt. We shuffled the human-composed sentences to induce variability in the generated sentences regarding participant style. Then, we divide the sentences into groups of 10 and use them as examples in the GPT-3 prompt, as shown in Figure \ref{fig:prompt_with_collected_sentences}. This prompt is inputted into the model, producing a completion following the same structure as the examples.
In Figure \ref{fig:prompt_with_collected_sentences}, we present the prompt used for sentence generation (a) and an English version (b) to facilitate reader understanding.
With this prompt, we generated 2,772 sentences, which are available for download\footnote{\url{https://drive.google.com/file/d/1gD8D9W-pEYuxgrCZK-jATh-v0rN_FCDY/view?usp=sharing}}.

\begin{figure}[t]
    \centering
    \footnotesize
\begin{subfigure}[b]{0.45\textwidth}
\begin{framed}
    
This is a list of distinct Portuguese sentences in direct order:
\newline
\newline
Example 1: eu brinquei de esconde-esconde com meus coleguinhas. 
\newline
\newline
Example 2: eu quero comer cuscuz. 
\newline
\newline
Example 3: eu gosto de ler muito. 
\newline
\newline
Example 4: o menino me bateu. 
\newline
\newline
Example 5: eu quero comer carne. 
\newline
\newline
Example 6: minha mãe fez comigo. 
\newline
\newline
Example 7: vamos voltar pra casa? 
\newline
\newline
Example 8: trocar a bombona de água. 
\newline
\newline
Example 9: eu brinquei com Maria ontem. 
\newline
\newline
Example 10: eu sou joão. 
\newline
\newline
Example 11:
\end{framed}
\caption{Used prompt.}
\end{subfigure}
\begin{subfigure}[b]{0.45\textwidth}
\begin{framed}
    
This is a list of distinct Portuguese sentences in direct order:
\newline
\newline
Example 1: I played hide and seek with my classmates.
\newline
\newline
Example 2: I want to eat couscous.
\newline
\newline
Example 3: I like to read a lot.
\newline
\newline
Example 4: The boy hit me.
\newline
\newline
Example 5: I want to eat meat.
\newline
\newline
Example 6: My mom did it to me.
\newline
\newline
Example 7: Shall we go home? 
\newline
\newline
Example 8: Change the water bottle. 
\newline
\newline
Example 9: I played with Maria yesterday. 
\newline
\newline
Example 10: I am John. 
\newline
\newline
Example 11:
\end{framed}
\caption{English version.}
\end{subfigure}
    \caption{GPT-3 text prompt for sentence generation based on examples of human-composed sentences.}
    \label{fig:prompt_with_collected_sentences}
\end{figure}{}

We used the words related to the pictogram in the Brazilian Portuguese subset of ARASAAC as a basis for generating new sentences through GPT-3. This vocabulary consists of 12,785 pictograms with words and expressions (e.g., \textquote{good morning}). Each pictogram has a list of keywords. In total, are 11,806 unique terms, including words and MWEs. We shuffled the vocabulary items and divided them into groups of 20. We randomly selected five words (or expressions) from each group and used them to search for example sentences on our already collected corpus. We sampled from 3 to 6 example sentences for each group and used them as few-shot examples on the GPT-3 prompt, as shown in Figure \ref{fig:prompt_with_vocabulary}.

\begin{figure}
    \centering
    \footnotesize
    \begin{subfigure}[b]{0.45\textwidth}
    \begin{framed}
        \medskip
These are examples of Portuguese sentences using the words in this vocabulary: \say{delas}, \say{vizinho}, \say{avó}, \say{médico}, \say{bebê}, \say{pai}, \say{professor}, \say{policial}, \say{garota}, \say{profissões}, \say{primas}, \say{irmã}, \say{crianças}, \say{rapaz}, \say{avô}, \say{de vocês}, \say{motorista}, \say{filho}, \say{dentista}, \say{adulto}.
\newline
\newline
\newline
Example 1: eu tenho um filho e uma filha. 
\newline
\newline
Example 2: eu vi meu filho feliz. 
\newline
\newline
Example 3: nós gostamos delas. 
\newline
\newline
Example 4: meu avô foi trabalhar. 
\newline
\newline
Example 5: você é um grande professor. 
\newline
\newline
Example 6: nós vamos seguir o professor. 
\newline
\newline
\newline
Example 7:

        \medskip
    \end{framed}
    \caption{Used prompt.}
    \end{subfigure}
    \begin{subfigure}[b]{0.45\textwidth}
    \begin{framed}
        \medskip
These are examples of Portuguese sentences using the words in this vocabulary: \say{their}, \say{neighbor}, \say{grandmother}, \say{doctor}, \say{baby}, \say{father}, \say{teacher}, \say{policeman}, \say{girl}, \say{professions}, \say{cousins}, \say{sister}, \say{children}, \say{boy}, \say{grandfather}, \say{from you}, \say{driver}, \say{son}, \say{dentist}, \say{adult}.
\newline
\newline
Example 1: I have a son and a daughter.
\newline
\newline
Example 2: I saw my son happy.
\newline
\newline
Example 3: we like them.
\newline
\newline
Example 4: my grandfather went to work.
\newline
\newline
Example 5: You are a great teacher.
\newline
\newline
Example 6: we are going to follow the professor.
\newline
\newline
Example 7:
        \medskip
    \end{framed}
    \caption{English version.}
    \end{subfigure}
    \caption{GPT-3 text prompt based on a controlled vocabulary for sentence generation.}
    \label{fig:prompt_with_vocabulary}
\end{figure}{}

\subsubsection{Data Cleaning} \label{sec:data_cleaning}

An automatically generated corpus like the one we produced can have misleading sentences. Therefore, we performed a data cleaning step, which consists of (a) removing sentences with offensive content using the method proposed by \cite{leite-etal-2020-toxic}; (b) removing sentences with higher perplexities according to BERTimbau \cite{souza2020bertimbau} and choosing the sentences in the first quartile for removal, and (c) removing sentences with less than three or more than 11 tokens. As mentioned in Section \ref{sec:perplexity}, BERT-like models do not directly compute perplexity for the cross-entropy is calculated only for masked tokens. However, if no masked token is inputted into the model and a copy of the input sentence is used as labels, it can assign a probability to each word in the sentence. We can then use this sentence probability to calculate the cross-entropy and the perplexity.

\subsubsection{Text to Pictogram} \label{sec:text_to_pictogram}

This section details how we transformed natural language sentences into pictograms. We used the Brazilian Portuguese set of pictograms from the ARASAAAC database. As mentioned before, each pictogram has a list of keywords, and each keyword has a glossary definition. However, a single term can be used for multiple pictograms. For example, the word \say{banco} (i.e., bank) has at least three pictograms. Thus, it is a word sense disambiguation problem. We solve this problem using BERTimbau \citep{souza2020bertimbau} to encode the target sentence and pictogram definitions and the K-Nearest Neighbor algorithm to choose the most relevant pictogram for each word in a given sentence.

\begin{figure}[t]
    \centering
    \begin{subfigure}[b]{0.15\textwidth}
        \centering
        \includegraphics[width=\textwidth]{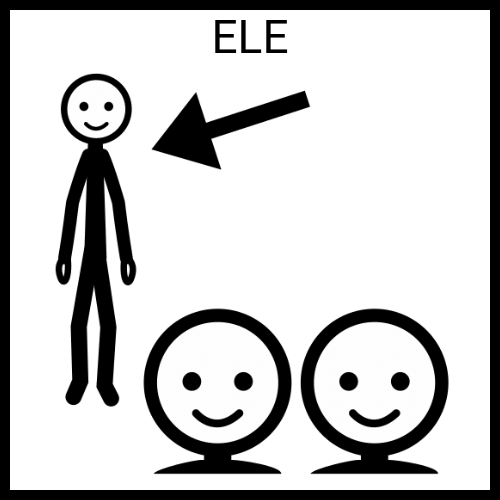}
    \end{subfigure}
    \begin{subfigure}[b]{0.15\textwidth}
        \centering
        \includegraphics[width=\textwidth]{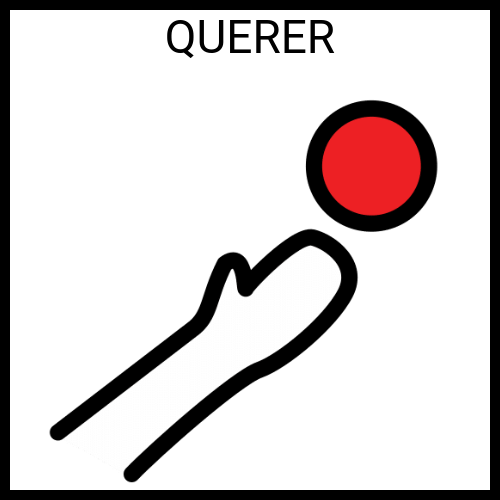}
    \end{subfigure}
    \begin{subfigure}[b]{0.15\textwidth}
        \centering
        \includegraphics[width=\textwidth]{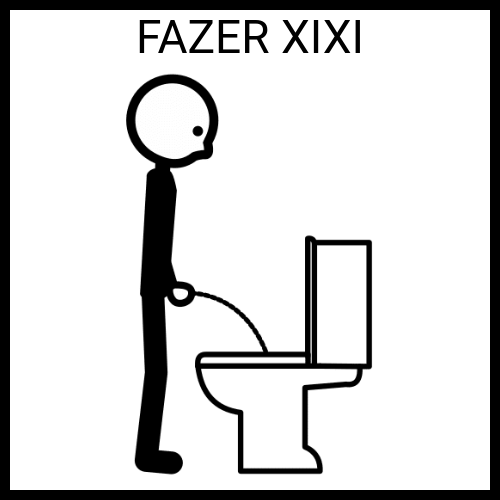}
    \end{subfigure}
    
    \caption{The sentence \textit{Ele quer fazer xixi} (he wants to pee) represented using ARASAAC pictograms.}
    \label{fig:text_to_pictogram}
\end{figure}

For example, given the sentence \say{ele quer fazer xixi} (he wants to pee), the first step is to tokenize it. We use all the keywords in ARASAAC as our vocabulary. It includes multi-word expressions (MWE) like \say{fazer xixi} (pee) or \say{café da manhã} (breakfast). For handling such expressions, we used a multi-word expression tokenizer\footnote{\url{https://www.nltk.org/_modules/nltk/tokenize/mwe.html}} in such a way that the tokenized version of the example sentence is $S_t = \{ele,\ querer,\ fazer\ xixi\}$. We also lemmatize the sentence, for the pictograms in ARASAAC have lemmas as keywords. We search the ARASAAC database for matching pictograms for each token in the sentence. When more than one pictogram is found, disambiguation is necessary. We concatenate the pictogram definitions and encode them using BERTimbau. We consider the sum of the hidden states of the last four encoder layers for the token [CLS] as the pictogram representation in an approach similar to \cite{scarlini2020ares}. For the target token, we consider as representation the vector that is the token position given the target sentence. In cases of MWEs, we consider the mean representation of the tokens in the expression. The final step is to get the pictogram representation most similar to the target representation using the KNN algorithm. Figure \ref{fig:text_to_pictogram} presents the pictogram version of the example sentence. For some words, there is no equivalent pictogram in ARASAAC. Still, we keep the word in the sentence considering that one can use a customized picture the represent it or a pictogram from another dataset.

\subsubsection{The constructed corpus} \label{sec:corpus_pt_analysis}

Table \ref{tab:pt_dataset_summary} summarizes the constructed corpus. The corpus consists of a set of 13,796 sentences that have the following characteristics: 1) are in direct order (i.e., subject+verb+complements); 2) are examples of phrases spoken in a conversation; 3) have a simple vocabulary; and 4) are common in the AAC context.

\input{tables/pt_dataset_summary}

\begin{figure}[t]
    \centering
    \begin{subfigure}[b]{0.45\textwidth}
        \centering
        \includegraphics[width=\textwidth]{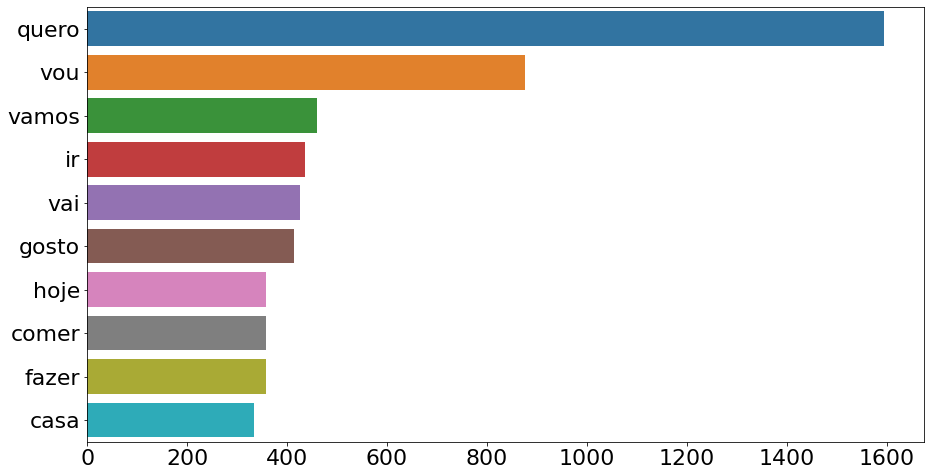}
        \caption{Words frequency.}
    \end{subfigure}
    \begin{subfigure}[b]{0.45\textwidth}
        \centering
        \includegraphics[width=\textwidth]{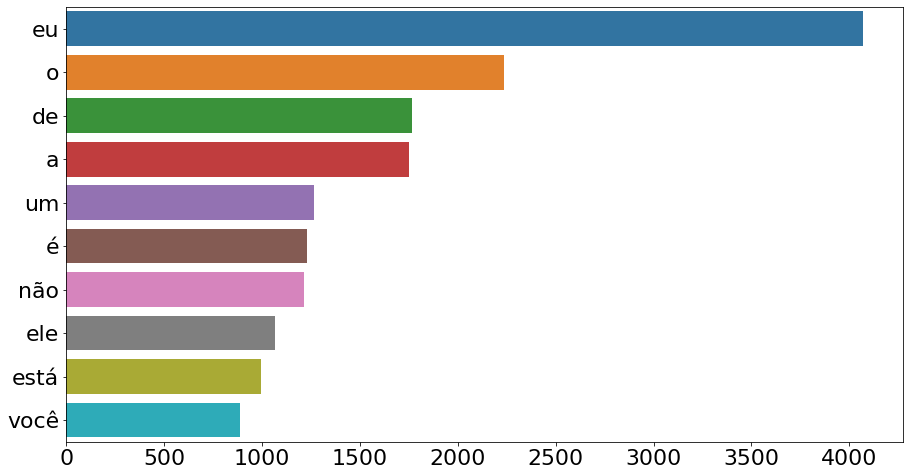}
        \caption{Stop-words frequency.}
    \end{subfigure}
    \caption{Frequency distribution of words in the constructed corpus.}
    \label{fig:word_frequency}
\end{figure}

Figure \ref{fig:word_frequency} presents a chart that displays the frequency of words in the corpus, with a separate section for stop words, sorted by frequency. The chart provides an overview of the most common terms used in the corpus. It can help identify patterns or trends in the language used. Notably, the most frequent word (excluding stopwords) in the corpus is \say{quero} (i.e., "I want"), suggesting a prominent focus on expressing wants or desires within the dataset. This aligns with the common usage of AAC systems, where users often communicate their needs and preferences. The high frequency of the word \say{quero} signifies a recurring theme of expressing intentions and personal desires in the sentences constructed by AAC practitioners and generated by GPT-3. 
The chart also displays the frequency of stop words, which are words that are not semantically meaningful, such as \say{o}, \say{a}, \say{de}, etc. 
Stop words in high frequency indicate that the corpus contains many common, everyday languages rather than specialized or technical ones. Overall, the chart in Figure \ref{fig:word_frequency} can be a useful tool for analyzing the language used in a corpus and gaining insight into the topics and themes it covers.

\begin{figure}[b]
    \centering
    \begin{subfigure}[b]{0.43\textwidth}
        \centering
        \includegraphics[width=\textwidth]{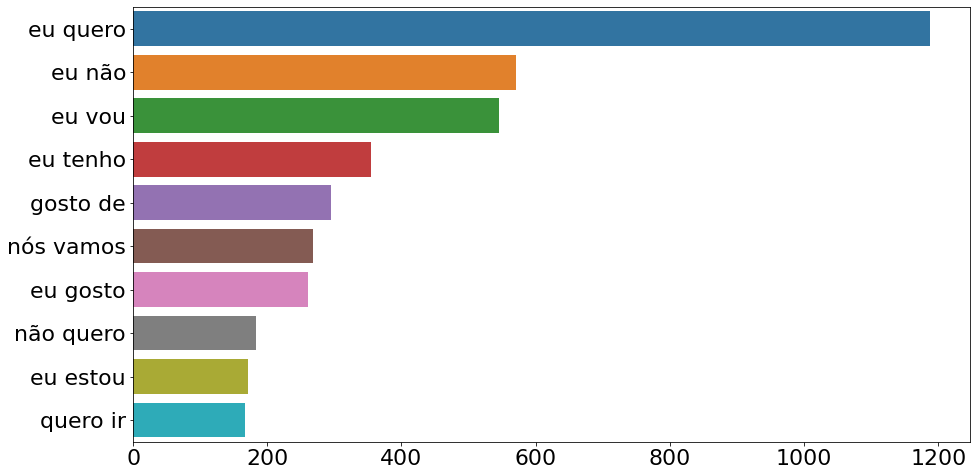}
        \caption{Bigram frequency.}
    \end{subfigure}
    \begin{subfigure}[b]{0.47\textwidth}
        \centering
        \includegraphics[width=\textwidth]{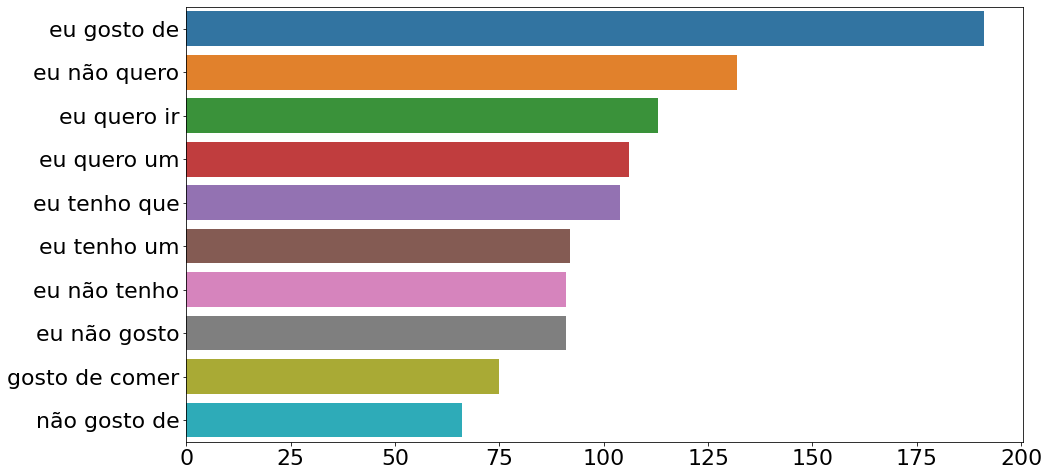}
        \caption{Trigram frequency.}
    \end{subfigure}
    \caption{Frequency distribution of N-gram in the constructed corpus.}
    \label{fig:ngram_frequency}
\end{figure}

The chart in Figure \ref{fig:ngram_frequency} displays the frequency of word combinations, specifically bigrams and trigrams, in the corpus. Bigrams are combinations of two words (e.g., \say{I am}), and trigrams are combinations of three words (e.g., \say{I am going}). The chart is sorted by frequency, with the most frequent bigrams and trigrams appearing at the top.
This type of analysis is useful for identifying common phrases and idiomatic expressions used in the corpus and understanding the relationship between words in the language. Additionally, it can provide insight into the style and tone of the text, such as whether it is formal or informal. Overall, the chart in Figure \ref{fig:ngram_frequency} can be a valuable tool for understanding the language used in the corpus at a deeper level. For example, the most frequent bigram is \say{eu quero} (I want), indicating that the corpus might be focused on expressing wants or desires. Additionally, it can be used to identify patterns in the language, such as specific conjunctions or prepositions, which can further inform the analysis of the corpus.

\begin{figure}[t]
    \centering
    \begin{subfigure}[b]{0.45\textwidth}
        \centering
        \includegraphics[width=\textwidth]{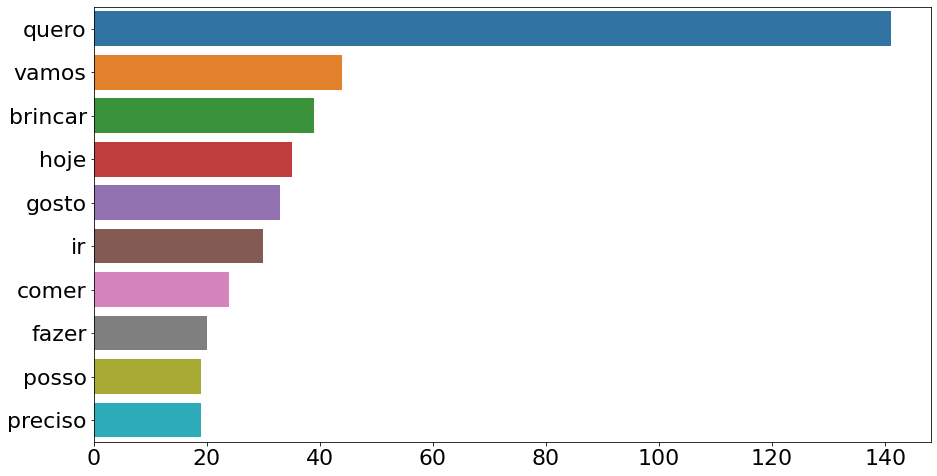}
        \caption{Words frequency.}
    \end{subfigure}
    \begin{subfigure}[b]{0.45\textwidth}
        \centering
        \includegraphics[width=\textwidth]{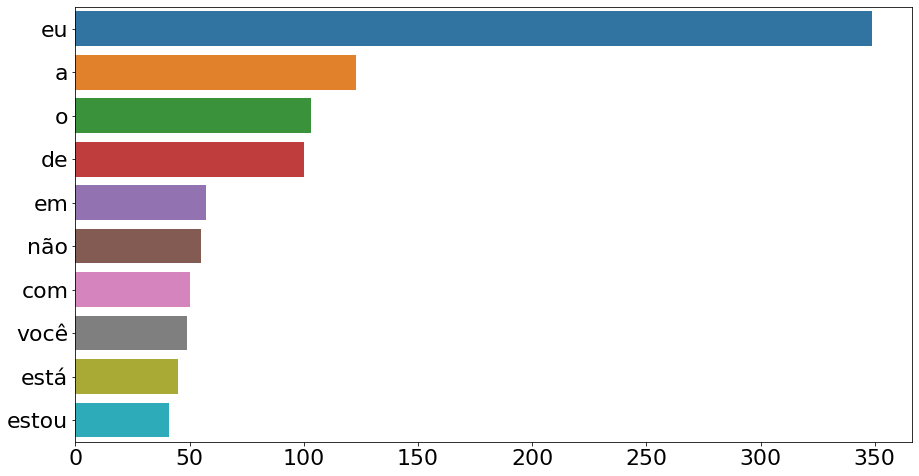}
        \caption{Stop-words frequency.}
    \end{subfigure}

    \begin{subfigure}[b]{0.45\textwidth}
        \centering
        \includegraphics[width=\textwidth]{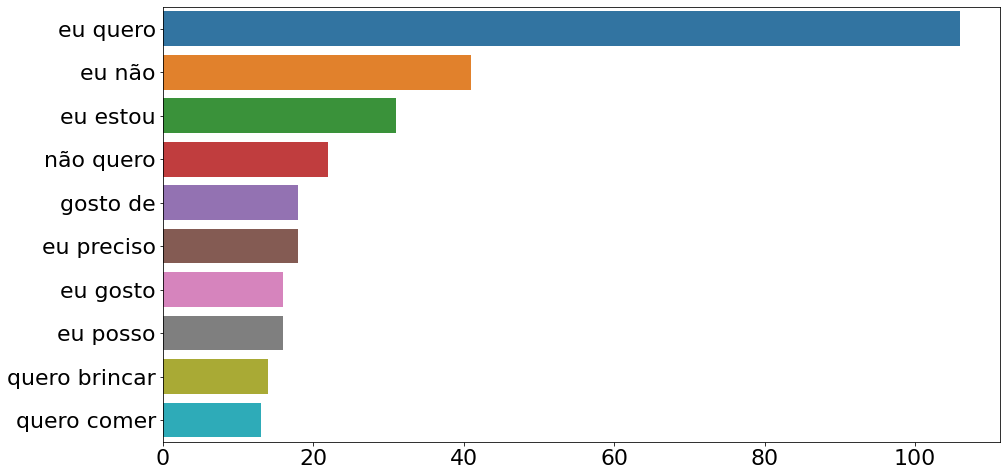}
        \caption{Bigram frequency.}
    \end{subfigure}
    \begin{subfigure}[b]{0.45\textwidth}
        \centering
        \includegraphics[width=\textwidth]{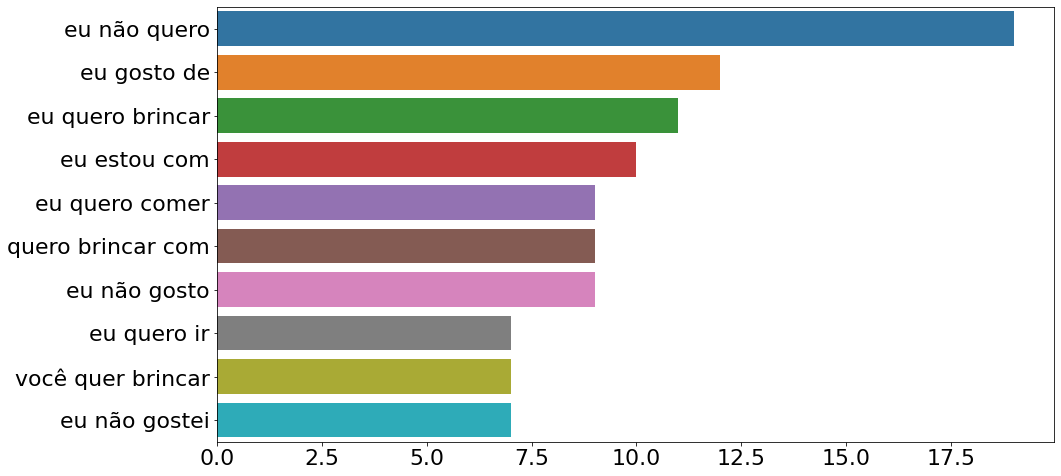}
        \caption{Trigram frequency.}
    \end{subfigure}
    \caption{Word and n-gram frequency distribution in the human-composed corpus.}
    \label{fig:word_frequency_human}
\end{figure}

Figure \ref{fig:word_frequency_human} presents the word and word combination (bigrams and trigrams) frequency distributions for the human-composed corpus. This figure provides a valuable basis for comparing the distribution of the generated corpus with the human-composed one. Upon analyzing the chart and comparing it to the frequency distributions shown in Figures \ref{fig:word_frequency} and \ref{fig:ngram_frequency}, it becomes evident that the human-composed and generated corpora exhibit similar patterns. Precisely, the frequency distribution of the most common words and stop words in the human-composed corpus aligns closely with their presence in the generated corpus. This similarity reinforces the effectiveness of using GPT-3 to generate synthetic sentences that resemble those composed by AAC practitioners. It indicates that the generated corpus captures essential linguistic patterns present in real-world AAC communication. The presence of similar word combinations (bigrams and trigrams) further corroborates the compatibility between the human-composed and generated corpora, strengthening the case for the effectiveness of the proposed methodology in creating a synthetic AAC dataset that mirrors the language patterns observed in real-life AAC interactions.

In addition to evaluating the quality and representativeness of the automatically generated sentences, we conducted a coverage assessment for the constructed corpus. The coverage measures the fraction of sentences generated through text augmentation that are assigned to the same cluster as at least one human-composed sentence. To quantify the coverage, we adopted a clustering-based approach, generating sentence embeddings for both the human-composed and augmented corpora. The k-means clustering algorithm was utilized to group the sentence embeddings into distinct clusters. For generating the sentence embeddings, we employed BERTimbau \cite{souza2020bertimbau}, using the average vector output from the last four encoder layers to represent the $[CLS]$ token. This methodology allowed us to effectively assess the degree of overlap between the human-composed and automatically generated sentences, shedding light on the corpus's overall coverage and ability to capture essential linguistic patterns.

To evaluate the coverage of the generated corpus, we collected an additional 203 sentences from AAC specialists. This set is referred to as the test set of the human-composed corpus. The original 667 sentences collected from the specialists constitute the training set of the human-composed corpus. The test set provides a means of measuring the quality and reliability of the generated corpus by comparing its content with the human-composed sentences.

\begin{figure}[t]
    \centering
    \centerline{\includegraphics[width=0.7\textwidth]{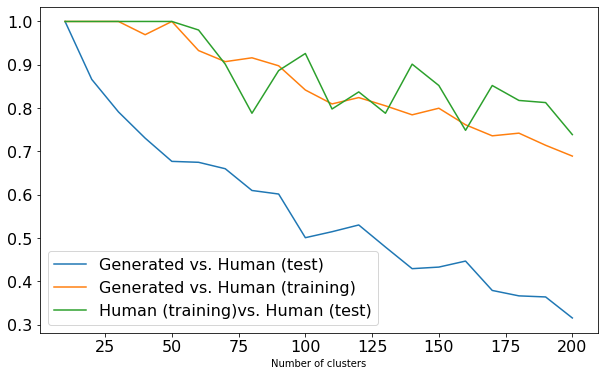}}
    \caption{Coverage of the automatically generated corpus over the human-composed sentences.}
    \label{fig:coverage}
\end{figure}

The line chart in Figure \ref{fig:coverage} depicts the coverage ratio of three different scenarios: the blue line represents the coverage ratio of the automatically generated corpus over the test set of the human-composed corpus. The orange line represents the automatically generated corpus coverage ratio over the human-composed corpus training set. Finally, the green line represents the coverage ratio of the test set of the human-composed corpus over the training set.

As the number of clusters increases from 10 to 200, we can observe that the blue line (coverage of the automatically-generated corpus over the test set of the human-composed corpus) decreases deeper than the other two lines. This can be explained by the fact that the human-composed corpus is smaller than the generated one, leading to a decrease in coverage as the number of clusters increases.
However, it is important to note that both the orange and green lines remain relatively stable throughout the range of the number of clusters, showing that the coverage of the auto-generated corpus over the training set and the test set of the human-composed corpus over the training set, respectively, is not significantly affected by the number of clusters.

The results demonstrate that the generated corpus is semantically similar to the original human-composed corpus, with a coverage ratio of up to 0.7 for the training set of the human-composed corpus, even when a large number of clusters is used. The coverage ratio is slightly lower but still significant for the test set of the human-composed corpus, remaining up to 0.5 with fewer than 130 clusters utilized.

\subsection{Fine-tuning} \label{sec:finetuning}

For fine-tuning BERTimbau for pictogram prediction, first, we have to change the model vocabulary and the input embeddings layer. BERT and BERTimbau use a vocabulary based on WordPiece \citep{Wu2016wordpiece}, which divides words into a limited set of common sub-word units (e.g., "Playing" into "Play\#" and "\#ing"). This vocabulary does not apply to pictogram prediction, for the tokens in pictogram-based sentences must be unique identifiers that cannot be divided into sub-items. For example, the sentence in Figure \ref{fig:text_to_pictogram} is represented as \say{6481 31141 16713}. Our vocabulary consists of identifiers for ARASAAC pictograms. This way, we use a word-level tokenizer, which splits words in a sentence by white spaces. 

Changing the vocabulary requires changing the embeddings layer, also. Intuitively, we tell the model that we changed the vocabulary to use a new language, and the new embedding vectors represent the words in this new language \citep{PEREIRA2022pictobert}. As mentioned in Section \ref{sec:method_inputs}, we use different approaches for pictogram embeddings. 

We use the corpus constructed with the method presented in Section \ref{sec:corpus_construction} as the training data. The corpus has a total of 13796 sentences, which we randomly divide with a proportion of 68/16/16 for training,  test, and validation sets. 
We fine-tune with a batch size of 768 sequences with 13 tokens (768 * 13 = 9,984 tokens/batch). Each data batch was collated to choose 15\% of the tokens for prediction, following the same rules as BERT: If the $i$-th token is chosen, it is replaced with 1) the $[MASK]$ token 80\% of the time, 2) a random token 10\% of the time, or 3) the unchanged $i$-th token 10\% of the time.
We use the same optimizer as BERT \citep{devlin-etal-2019-bert}: Adam, with
a learning rate of $1 \times 10^{-5}$ for all model versions, with $\beta_1 = 0.9$, $\beta_2 = 0.999$, L2 weight decay of 0.01, and linear decay of learning rate. Fine-tuning was performed in a single 16GB NVIDIA Tesla V100 GPU for 200 epochs for the captions and synonyms versions and 500 epochs for the other versions. The definition- and image-based versions require more training time because the input vectors are from a different vectorial space than the original BERTimbau embeddings.

%% file: tables/pt_dataset_summary.tex
\begin{table}[b]
\centering
\footnotesize
\caption{Portuguese dataset summary.}
\label{tab:pt_dataset_summary}
\begin{tabular}{@{}cc|ccccc@{}}
\toprule
\multicolumn{2}{c|}{\textbf{Words}} &  \multicolumn{5}{c}{\textbf{Sentences}}                  \\ \midrule
\textbf{Total}  & \textbf{Unique}  & \textbf{Total} & \textbf{Max Length} & \textbf{Min Length} & \textbf{Mean Length} & \textbf{Most Frequent Length} \\
\midrule
89572           & 4758             & 13796           & 11                  & 3                   & 6                    & 6 (3432 times)                \\ \bottomrule
\end{tabular}
\end{table}

%% file: sections/results.tex
\section{Results and Analysis} \label{sec:results}

Table \ref{tab:results} presents the results obtained by testing each version of the proposed model regarding perplexity (PPL) and Top-$n$ accuracies. We use $n \in \{1,9,18,25,36\}$ to simulate the different grid sizes an AAC system can have. We calculate perplexity by exponentiating the cross-entropy over the test set's entire sentences without masked tokens (cf. Section \ref{sec:perplexity}). For perplexity, lower is better. The table shows that the model in which the embeddings were calculated using the pictogram caption synonyms has the lowest perplexity. This means that this model better understands how the language present in the test set works. Seeing new data than the other model versions was intuitively less surprising. Thus, it can perform a better generalization in different scenarios. The model with embeddings extracted only from pictograms' captions had better accuracy. However, the difference between these two models in all metrics is not as significant enough to indicate which is the best. 

\input{tables/results.tex}

Regarding the models in which the pictogram definitions were used to compute embeddings, using the mean vector of the definition extracted from the BERTimbau input embeddings was shown to be more effective. Using the BERTimbau outputs as the definition representation did not show good results, with higher perplexities and lower accuracies.  Fine-tuning BERTimbau using these embeddings may require more training data and time, for the vectors are from a vectorial space different from the model's original. The same happens to the models using embeddings computed from pictogram images and their combinations. Based on these models' training and validation loss curves, there is still space for improvement, as the measures keep falling even after 500 epochs.

Therefore, based on the metrics presented in Table \ref{tab:results}, the best way to represent a pictogram in the proposed method is using the pictogram caption or its synonyms. The decision of which of these two approaches to use depends on the vocabulary characteristics. For example, it is impossible to use synonyms if no synonyms dataset is available. However, if the same caption is used for two different pictograms in a vocabulary, it may be difficult for the model to disambiguate them. Using the pictogram concept, as in \citep{PEREIRA2022pictobert}, can solve these problems. Nevertheless, to our knowledge, there is not for well-established Brazilian Portuguese lexical database as the Princeton WordNET is for English. Using the pictogram definition can be an alternative, but the results demonstrate that it performs worse than using only captions or synonyms. In addition, encoding pictograms based on their definition may require more time and resources than using captions or related words.

Figure \ref{fig:generated_examples} presents four sentences from the test dataset and the top-6 predictions performed by the model trained using embeddings from captions. The examples show the model behavior in different scenarios. 
Figure \ref{fig:example1} presents an example of a subject+verb+complement sentence. The sentence represented is equivalent to \say{you want a \rule{1cm}{0.15mm}.} or \say{do you want a \rule{1cm}{0.15mm}?} in English. Thus, it can be an affirmation or a question in construction. The top-6 pictogram suggested as completions demonstrate that the model prediction is affected by the token \textit{um} (or \textit{a}), which is a preposition. 
Figure \ref{fig:example2} presents an example using an auxiliary verb (i.e., \textit{ir}, to go). In this case, the model predicts pictograms that can act as the sentence's main verb.
Figure \ref{fig:example3} shows an example of descriptor prediction. Notice that there are two pictograms to the word \textit{novo} (i.e., new). It occurs because they have the same caption. 
Figure \ref{fig:example4} presents an example of predicting the second pictogram of a sentence that begins with a verb. In this case, the verb \textit{estar} can mean \textit{I am} (e.g., I am tired) or \textit{it is} (e.g., It is hot).

\newcommand{\rulesep}{\unskip\ \vrule\ }
\begin{figure}[]
    \centering
    \begin{subfigure}[b]{0.45\textwidth}
        \centering
        \includegraphics[width=\textwidth,trim={0 3cm  0 0}]{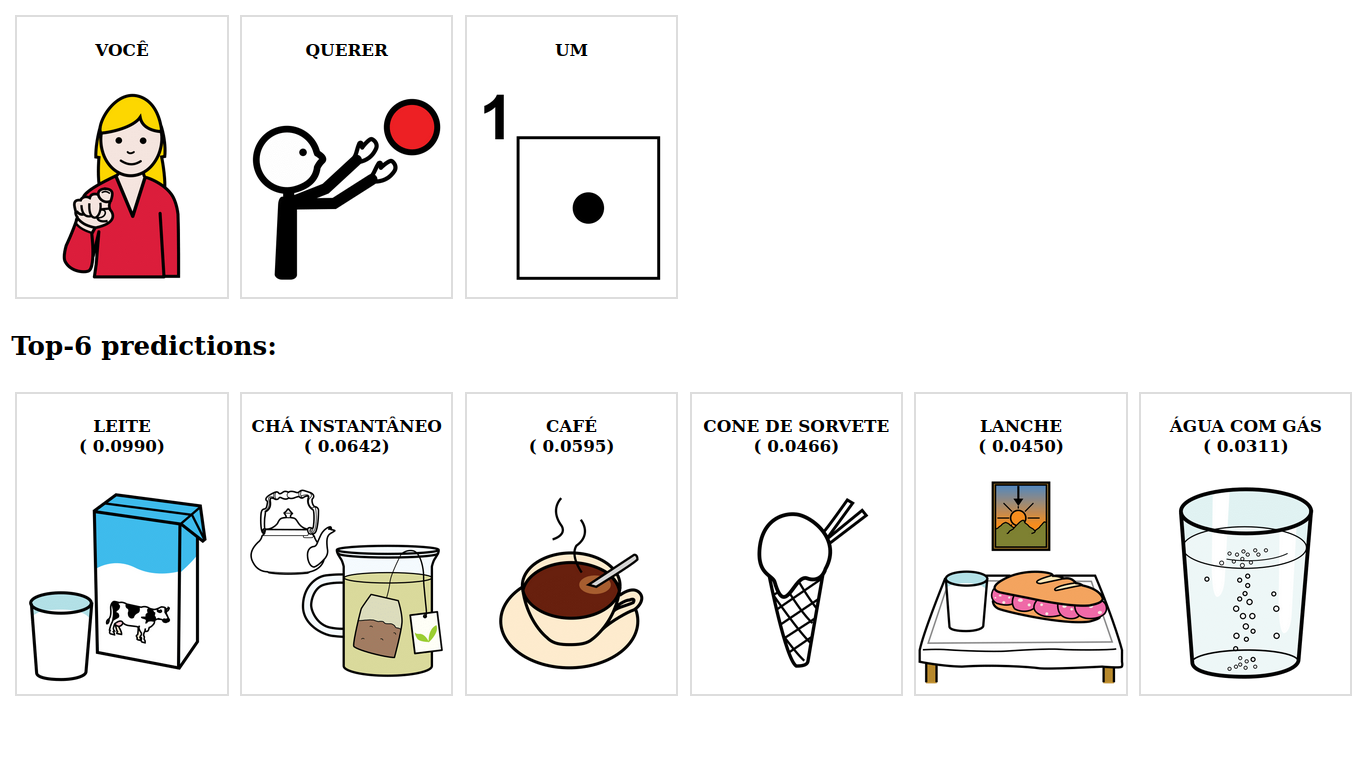}
        \caption{}
        \label{fig:example1}
    \end{subfigure}
    \rulesep
    \begin{subfigure}[b]{0.45\textwidth}
        \centering
        \includegraphics[width=\textwidth,trim={0 3cm  0 0}]{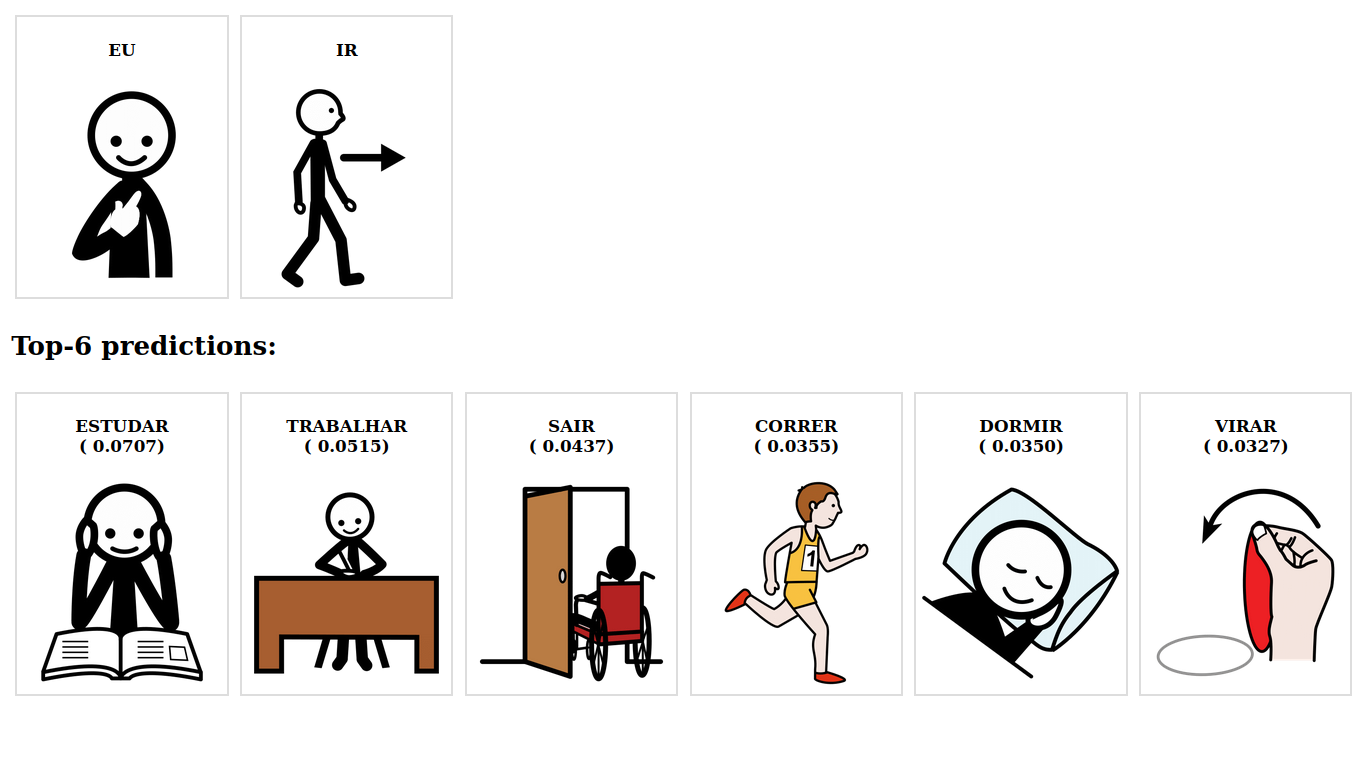}
        \caption{}
        \label{fig:example2}
    \end{subfigure}
    \begin{subfigure}[b]{0.45\textwidth}
        \centering
        \includegraphics[width=\textwidth,trim={0 3cm  0 0}]{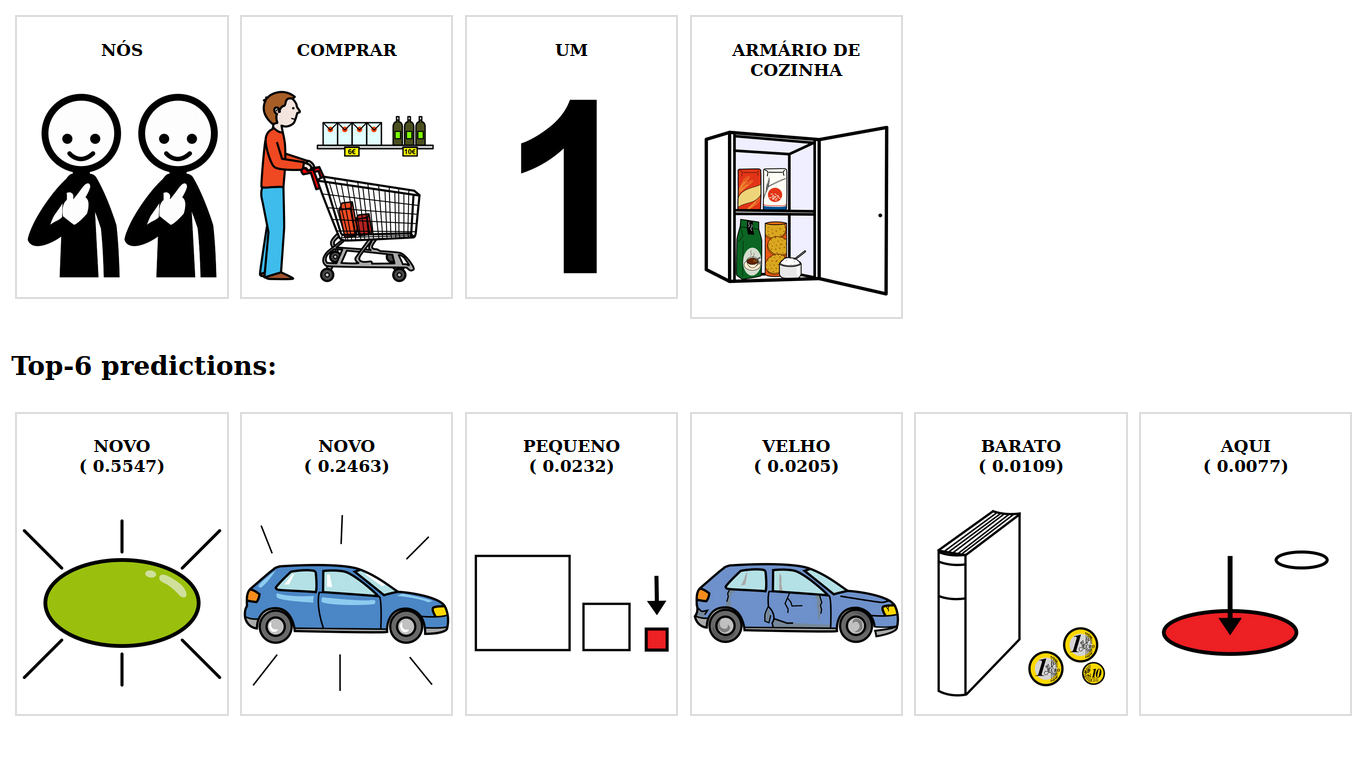}
        \caption{}
        \label{fig:example3}
    \end{subfigure}
    \rulesep
    \begin{subfigure}[b]{0.45\textwidth}
        \centering
        \includegraphics[width=\textwidth,trim={0 3cm  0 0}]{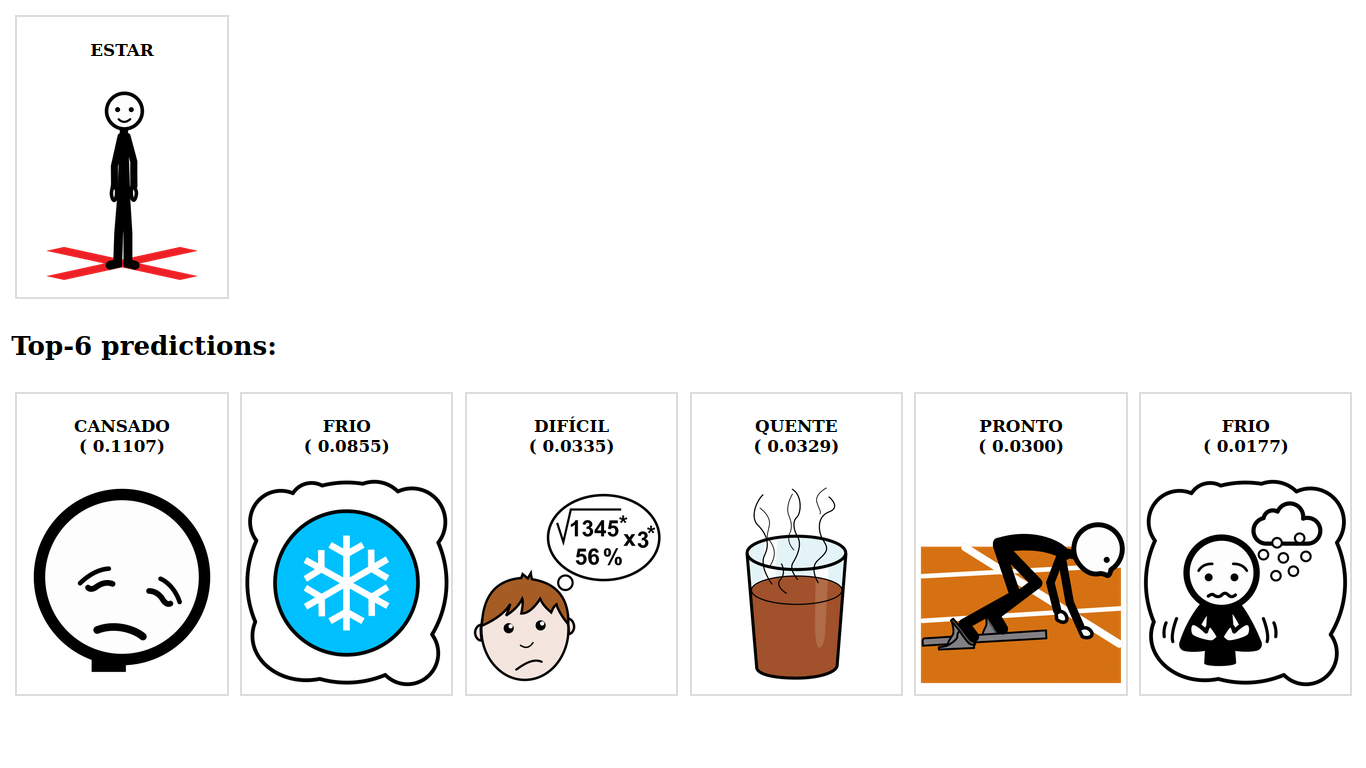}
        \caption{}
        \label{fig:example4}
    \end{subfigure}
    
    \caption{Example of predictions made by the captions model.}
    \label{fig:generated_examples}
\end{figure}

\subsection{Usage Guidelines: How Can Others Use This Work?}

Researchers, developers, and practitioners interested in utilizing the proposed method and findings presented in this work can follow the guidelines outlined below to enhance pictogram prediction in AAC systems, considering it as a low-resource domain:

\begin{itemize}
    \item \textbf{Constructing a Synthetic AAC Corpus}: Researchers can extend the method for constructing the synthetic AAC corpus to create their corpus. This approach can be applied to different languages or specific target populations. By following the methodology described in Section \ref{sec:corpus_construction}, researchers can adapt the process and gather data relevant to their specific context and objectives. It is worth mentioning that the generated corpus depends on the input sentences and vocabulary. Furthermore, it is possible to induce the model to generate sentences of a specific context or user or groups of user needs. This allows for a more tailored approach to creating a synthetic AAC corpus catering to specific requirements or preferences.
    \item \textbf{Fine-tuning a Language Model}: The constructed synthetic AAC corpus can be used for fine-tuning transformer-based language models such as BERT. Researchers can combine the corpus with the methodology presented in Section \ref{sec:finetuning} to adapt the language model for pictogram prediction. This process allows for personalized message authoring in AAC systems, enhancing the system's relevance and accuracy in generating suggestions.
    \item  \textbf{How to represent a pictogram}: Researchers or AAC developers can use our experiments as a basis to decide how to represent a pictogram when using a transformer-based model. Our experiments have shown that when it comes to representing pictograms, there are a few different approaches that yield similar results. One way is to use the captions associated with the pictograms, treating the prediction task as a word prediction task. However, it is important to consider that in some vocabularies different pictograms can have the same caption having the same vectorial representation. Another approach is to use synonyms or definitions, but this requires access to an external database that may not always be available. These findings may be helpful for developers and researchers looking to work with pictogram prediction.
    \item \textbf{Developing AAC Systems with Pictogram Prediction}: Developers can leverage the proposed method to design AAC systems that perform pictogram word prediction based on the user's vocabulary. To implement this, developers can utilize the method we proposed to create a corpus, modify a transformer-based model, and train it accordingly. By incorporating pictogram prediction, AAC systems can enhance the user experience, providing real-time suggestions that facilitate efficient and effective communication.
\end{itemize}

\subsection{Limitations}

The limitations of this study primarily stem from the fact that the proposed models were not evaluated in a real-world AAC system by actual users, either with or without CCN. This is a significant limitation as the effectiveness and efficiency of AAC solutions can be best evaluated in a practical setting, where users interact with the system in their daily communication. However, developing a fully functional AAC system that incorporates the models proposed in this paper is beyond the scope of this study. This study focused on developing and evaluating the models in a controlled environment, which may not fully reflect the complexities and challenges of real-world AAC usage.

Another limitation of this study is the lack of diversity in the AAC corpus used for training the model. The corpus was constructed using sentences generated by AAC practitioners and synthetic sentences generated by GPT-3, which may not fully represent the diverse communication needs and styles of AAC users. Nevertheless, it's important to note that constructing the corpus is a crucial step in our methodology that can be replicated for various scenarios. The output of corpus generation is dependent on the input sentences and vocabulary. If a diverse set of sentences is used, it may lead to a more varied corpus. However, we should also consider the costs associated with corpus generation, which can limit the quantity of generated sentences and ultimately affect the corpus's diversity.
Additionally, this study focused on evaluating the models' performance in Brazilian Portuguese and did not explore their applicability to other languages. The effectiveness of the models in different languages may vary due to language-specific characteristics and variations in vocabulary usage. Further research is needed to adapt and evaluate the models for other languages.

Furthermore, we also recognize as a limitation the fact that we used a model that was not specifically trained for Brazilian Portuguese for generating the sentences. This could affect the generated sentences' accuracy and relevance, as the model may not fully capture the nuances and complexities of the Brazilian Portuguese language. Future studies could consider using a model specifically for Brazilian Portuguese like Sabiá \cite{pires2023sabia}.

Finally, the study did not consider the potential influence of user-specific factors on the models' performance, such as age, cognitive abilities, or familiarity with AAC systems. These factors can significantly impact the usability and effectiveness of AAC systems. Future research should explore these factors to optimize the models for individual users and address their unique communication needs.

Despite these limitations, this study provides valuable insights into using BERT-like models for pictogram prediction in AAC systems and lays the groundwork for future research.

%% file: tables/results.tex
\begin{table}[b]
\centering
\footnotesize
\caption{Evaluation results by model version descending sorted by ACC@1. ACC@\{1, 9, 18, 25, 36\} simulate the different grid sizes an AAC system can have.}
\label{tab:results}
\begin{tabular}{@{}lcccccc@{}}
\toprule
Method                                           & PPL     & ACC@1 & ACC@9 & ACC@18 & ACC@25 & ACC@36 \\ \midrule
Pictogram captions                               & 15.433  & \textbf{0.237} & \textbf{0.530} & \textbf{0.620}  & \textbf{0.657}  & \textbf{0.702}  \\
Pictogram synonyms                               & \textbf{14.282}  & 0.225 & 0.511 & 0.604  & 0.647  & 0.698  \\
Pictogram definition {[}input embeddings mean{]} & 23.368  & 0.209 & 0.492 & 0.580  & 0.627  & 0.673  \\
Pictogram image + synonyms                       & 122.407 & 0.042 & 0.169 & 0.220  & 0.255  & 0.293  \\
Pictogram definition {[}mean last layer{]}       & 22.496  & 0.019 & 0.122 & 0.206  & 0.246  & 0.295  \\
Pictogram image                                  & 106.130 & 0.007 & 0.037 & 0.078  & 0.112  & 0.146  \\
Pictogram image + caption                        & 89.685  & 0.007 & 0.038 & 0.076  & 0.111  & 0.146  \\
Pictogram definition {[}CLS{]} last layer        & 89.107  & 0.003 & 0.062 & 0.117  & 0.153  & 0.203  \\ \bottomrule
\end{tabular}
\end{table}

%% file: sections/conclusions.tex
\section{Conclusions} \label{sec:conclusions}

Recent studies propose methods for pictogram prediction in AAC systems as an alternative to support the construction of meaningful and grammatically adequate sentences. The existing methods vary regarding the technique used for prediction and how to represent a pictogram. In AAC, a pictogram (a.k.a. communication card) combines a pictograph and a caption representing a concept (e.g., an action, person, object, or location). Some studies consider that the word or expression in the caption is enough to perform prediction. At the same time, others prefer to represent the pictogram as a ser of synonyms or a concept from a dictionary.

In this paper, we investigate the most appropriate known manner to represent pictograms for pictogram prediction in Brazilian Portuguese. To do this, we propose a method for finetuning a BERT-like model for pictogram prediction from scratch that might be suitable for languages other than Portuguese. First, we constructed an AAC corpus for Portuguese by collecting sentences from specialists and augmenting the data using a large language model. Then, we finetuned BERTimbau, a Portuguese version of BERT. We conducted experiments using four different ways of representing pictograms: 1) using the captions (i.e., words or expressions), 2) using the caption synonyms, 3) using the pictogram definition, and 3) using the pictogram image to compute embeddings.
We evaluated the performance of the models in terms of perplexity and top-$n$ accuracy. The results demonstrate that using embeddings computed from the pictograms' caption, synonyms, or definitions have a similar performance. Using synonyms leads to lower perplexity, but using captions leads to the highest accuracies. This suggests that choosing a method to implement in an AAC system is a design decision. Additionally, we found that using image representations did not improve the quality of pictogram prediction. 

We recognize using a synthetic corpus as a limitation of this study. Although the corpus was constructed using human-composed sentences as a basis, the resulting sentences can suffer the influence of the GPT-3 training biases \citep{dale_2021}. To reduce this impact, we removed the sentences with offensive content. In addition, GPT-3 can generate incoherent sentences with confusing semantics as it was not trained specifically for Portuguese. This can also affect the diversity of produced sentences, as it must have seen less Portuguese text than English during training. Initiatives such as the Open Pre-trained Transformers (OPT) \citep{zhang2022opt} might boot the emergence of models trained in languages other than English, which can lead to more comprehensive and coherent text generation. An example is Sabiá \cite{pires2023sabia}, a model trained for Brazilian Portuguese that we intend to use in future works.
We also recognize that humans (with or without CCN) can assess AAC solutions more accurately. However, to do so, an AAC system using the proposed models for prediction is required, which is not addressed in this paper. We propose a method to train models to be plugged into end-to-end applications that consider the particular needs of each user or group of users.

In future work, we intend to evaluate the model prediction quality by testing it with AAC users' parents and caregivers and then with people with CCN. Besides, we intend to implement a text expansion system for Brazilian Portuguese capable of expanding telegraphic sentences (e.g., \textit{eu comer bolo escola ontem}, i.e., I eat cake school yesterday) to natural language expanded sentences (e.g., \textit{eu comi bolo na escola ontem}, i.e., I ate a cake at school yesterday). The text expansion might help the interlocutor understand what the AAC user says.